\documentclass[11pt,a4paper]{article}
\usepackage[hyperref]{acl2017}
\usepackage{times}
\usepackage{latexsym}
\usepackage{amsmath}
\usepackage{graphicx}
\usepackage{url}

\aclfinalcopy % Uncomment this line for the final submission
\title{Survey of Visual Question Answering: Datasets and Techniques}
\author{Akshay Kumar Gupta \\
Indian Institute of Technology Delhi\\
\texttt{cs5130275@cse.iitd.ac.in}}

\date{}

\begin{document}
\maketitle
\begin{abstract}
  Visual question answering (or VQA) is a new and exciting problem that combines
  natural language processing and computer vision techniques. We present 
  a survey of the various datasets and models that have been used to tackle 
  this task. The first part of this survey details the various datasets for VQA
  and compares them along some common factors. The second part of this survey details
  the different approaches for VQA, classified into four types: non-deep learning
  models, deep learning models without attention, deep learning models with
  attention, and other models which do not fit into the first three. Finally, 
  we compare the performances of these approaches and provide some directions for
  future work.
\end{abstract}

\section{Introduction}

Visual Question Answering is a task that has emerged in the last few years and has been getting a lot of attention from the machine learning community \cite{antol2015vqa} \cite{wu2016visual}. The task typically involves showing an image to a computer and asking a question about that image which the computer must answer. The answer could be in any of the following forms: a word, a phrase, a yes/no answer, choosing out of several possible answers, or a fill in the blank answer.

Visual question answering is an important and appealing task because it combines the fields of computer vision and natural language processing. Computer vision techniques must be used to understand the image and NLP techniques must be used to understand the question. Moreover, both must be combined to effectively answer the question in context of the image. This is challenging because historically both these fields have used distinct methods and models to solve their respective tasks.

This survey describes some prominent datasets and models that have been used to tackle the visual question answering task and provides a comparison on how well these models perform on the various datasets. Section 2 covers VQA datasets, Section 3 describes models and Section 4 discusses the results and provides some possible future directions.

\begin{table*}[]
\centering
\label{my-label}
\hspace*{-10mm}
\begin{tabular}{c|cccccc}
\textbf{}               & \textbf{\begin{tabular}[c]{@{}c@{}}Number of \\ Images\end{tabular}} & \textbf{\begin{tabular}[c]{@{}c@{}}Number of \\ Questions\end{tabular}} & \textbf{\begin{tabular}[c]{@{}c@{}}Avg. questions\\ per image\end{tabular}} & \textbf{\begin{tabular}[c]{@{}c@{}}Average question\\ length\end{tabular}} & \textbf{\begin{tabular}[c]{@{}c@{}}Average answer\\ length\end{tabular}} & \textbf{\begin{tabular}[c]{@{}c@{}}Q/A\\ generation\end{tabular}}\\ \hline
\textbf{DAQUAR}         & 1,449                                                                & 12,468                                                                  & 8.60                                                                        & 11.5                                                                       & 1.2          & Human                                                            \\
\textbf{Visual7W}       & 47,300                                                               & 327,939                                                                 & 6.93                                                                        & 6.9                                                                     & 2.0            & Human                                                          \\
\textbf{Visual Madlibs} & 10,738                                                               & 360,001                                                                 & 33.52                                                                       & 4.9                                                                       & 2.8           & Human                                                            \\
\textbf{COCO-QA}        & 117,684                                                              & 117,684                                                                 & 1.00                                                                        & 9.65                                                                       & 1.0          & Automatic                                                            \\
\textbf{FM-IQA}         & 158,392                                                              & 316,193                                                                 & 1.99                                                                        & 7.38 (Chinese)                                                             & 3.82 (Chinese)            & Human                                               \\
\textbf{VQA (COCO)}     & 204,721                                                              & 614,163                                                                 & 3.00                                                                        & 6.2                                                                        & 1.1         & Human                                                             \\
\textbf{VQA (Abstract)} & 50,000                                                               & 150,000                                                                 & 3.00                                                                        & 6.2                                                                        & 1.1         & Human                                                            
\end{tabular}
\caption{VQA Datasets}
\end{table*}

\section{Datasets}

Several large-scale datasets have been released in the past 2-3 years for the VQA task. We discuss these datasets below. Table 1 contains a summary of these datasets.

\subsection{DAQUAR \cite{malinowski2014multi}}
The DAtaset for QUestion Answering on Real-world images (or DAQUAR), released in 2015, was the first dataset and benchmark released for the VQA task. It takes images from the NYU-Depth V2 dataset which contains images along with their semantic segmentations. Every pixel of an image is labeled with an object class (or no object) out of 894 possible classes. The images are all of indoor scenes. There are a total of 1449 images (795 training, 654 test). The authors generated question answer pairs in two ways: 1) Automatically, using question templates. The authors define 9 templates for questions, whose answers can be taken from the existing NYU-Depth V2 dataset annotations. An example of a question template is `How many [object] are in [image id]?'. 2) Using human annotations. They asked 5 participants to generate questions and answers with the constraint that answers must be either colors, numbers or classes or sets of these. The resultant dataset contains a total of 12468 question-answer pairs (6794 training, 5674 test). A reduced dataset containing only 37 object classes is also available.

The authors propose two evaluation metrics for this dataset: One is simple accuracy, which is not a very good metric for multi-word answers, and the second is WUPS score which gives a score for a generated answer between 0.0 and 1.0 based on average match between answer and ground truth answers. Typically the WUPS score is thresholded at 0.9 (That is, if the WUPS score for an answer is above 0.9 then it is correct.)
\begin{figure}[h!]
\includegraphics{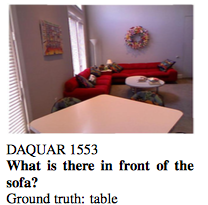}
\caption{Taken from \cite{ren2015exploring}}
\end{figure}
\subsection{Visual7W \cite{zhu2016visual7w}}
Visual 7W is a dataset generated using images from the MS-COCO dataset \cite{lin2014microsoft} for image captioning, recognition and segmentation. The Visual7W dataset gets it name from generating multiple-choice questions of the form (Who, What, Where, When, Why, How and Which). Workers on Amazon Mechanical Turk (AMT) were used to generate the questions. A separate set of three workers were used to rate the questions and those with less than two positive votes were discarded. Multiple choice answers were generated both automatically and by AMT workers. AMT workers were also asked to draw bounding boxes of objects mentioned in the question in the image, firstly to resolve textual ambiguity (Eg. An image has two red cars. Then `red car' in the question could refer to either of these.), and secondly to enable answers of a visual nature (`pointing' at an object). The dataset contains 47,300 images and 327,939 questions.

\subsection{Visual Madlibs \cite{yu2015visual}}
The Visual Madlibs dataset is a fill-in-the-blanks as well as multiple choice dataset. Images are collected from MS-COCO. Descriptive fill-in-the-blank questions are generated automatically using templates and object information. Each question generated in this way is answered by a group of 3 AMT workers. The answer can be a word or a phrase. Multiple choices for the blanks are also provided as an additional evaluation benchmark. The dataset contains 10,738 images and 360,001 questions. The multiple choice questions are evaluated on the accuracy metric.

\subsection{COCO-QA \cite{ren2015exploring}}
The COCO-QA dataset is another dataset based on MS-COCO. Both questions and answers are generated automatically using image captions from MS-COCO and broadly belong to four categories: Object, Number, Color and Location. There is one question per image and answers are single-word. The dataset contains a total of 123,287 images. Evaluation is done using either accuracy or WUPS score.
\begin{figure}[h!]
\includegraphics{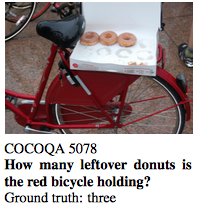}
\caption{Taken from \cite{ren2015exploring}}
\end{figure}
\subsection{FM-IQA \cite{gao2015you}}
The Freestyle Multilingual Image Question Answering dataset (FM-IQA) takes images from the MS-COCO dataset and uses the Baidu crowdsourcing server to get workers to generate questions and answers. Answers can be words, phrases or full sentences. Question/answer pairs are available in Chinese as well as their English translations. The dataset contains 158,392 images and 316,193 questions. They propose human evaluation through a visual Turing Test, which may be one reason this dataset has not gained much popularity.

\subsection{VQA \cite{antol2015vqa}}
The Visual Question Answering (VQA) dataset is the most widely used dataset for the VQA task. This dataset was released as part of the visual question answering challenge. It is divided into two parts: one dataset contains real-world images from MS-COCO, and another dataset contains abstract clipart scenes created from models of humans and animals to remove the need to process noisy images and only perform high level reasoning. Questions and answers are generated from crowd-sourced workers and 10 answers are obtained for each question from unique workers. Answers are typically a word or a short phrase. Approximately 40\% of the questions have a yes or no answer. For evaluation, both open-ended answer generation as well as multiple choice formats are available. Multiple choice questions have 18 candidate responses. To evaluated open-ended answers, a machine generated answer is normalized by the VQA evaluation system and then evaluated as Score = min(\#humans who provided that exact answer / 3, 1).  So, an answer is considered completely correct if it matches the responses of at least three human annotators. If it matches none of the 10 candidate responses then it is given a score of 0. The original VQA dataset has 204,721 MS-COCO images with 614,163 questions and 50,000 abstract images with 150,000 questions. The 2017 iteration of the VQA challenge has a bigger dataset with a total of 265,016  MS-COCO and abstract images and an average of 5.4 questions  per image. The exact number of questions is not mentioned on the challenge website.
\begin{figure}[h!]
\includegraphics{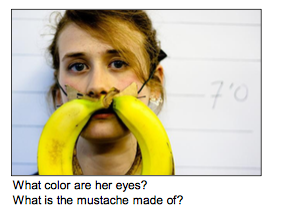}
\caption{Taken from \cite{antol2015vqa}}
\end{figure}
\section{Models}
The VQA task was proposed after deep learning approaches had already gained wide popularity due to their state-of-the-art performance on various vision and NLP tasks \cite{krizhevsky2012imagenet} \cite{bahdanau2014neural}. As a result almost all work on VQA in the literature involves deep learning approaches, as opposed to more classical approaches like graphical models. There are a couple of models which use a non-neural approach, which are detailed in the first subsection. In addition, several simple baselines that authors use involve non-neural methods, which are also described. The second sub-section describes deep learning models which do not involve the use of attention-based techniques. The third sub-section details attention-based deep learning models for VQA. Results of all the models described are summarized in Tables 2 and 3.
\subsection{Non-deep learning approaches}
\subsubsection{Answer Type Prediction (ATP)}
\cite{kafle2016answer} propose a Bayesian framework for VQA in which they predict the answer type for a question and use this to generate the answer. The possible answer types vary across the datasets they consider. For instance, for COCO-QA they consider four answer types: object, color, counting, and location.

Their model computes the probability of an answer $a$ and answer type $t$ given the image $x$ and question $q$,
\begin{equation}
\resizebox{0.5\textwidth}{!}{$P(A=a,T=t|x,q) = \frac{P(x|A=a,T=t,q)P(A=a|T=t,q)P(T=t|q)}{P(x|q)}$}
\end{equation}
which follows from Bayes' rule. They can then marginalize over all answer types to obtain $P(A=a|x,q)$. The denominator is constant for a given question and image and can thus be ignored.

They model the three probabilities in the numerator with three separate models. The second and third probabilities are both modeled using logistic regression. The features used for the question was the skip-thought vector representation \cite{kiros2015skip} of the question (They use the pre-trained skip thought model). The first probability is modeled as a conditional multivariate Gaussian, similar in principle to Quadratic Discriminant Analysis. The original image features are used in this model.

The authors also introduced some simple VQA baselines, like feeding only image features or only question features to a logistic regression classifier, feeding both image and question features to a logistic regressor, and feeding the same features to a multi-layer perceptron. They evaluated results on the DAQUAR, COCO-QA, Visual7W and VQA datasets.

\subsubsection{Multi-World QA \cite{malinowski2014multi}}
This paper models the probability of an answer given a question and an image as 
\begin{equation}
P(A=a|Q, W) = \Sigma_{T}P(A=a|T,W)P(T|Q)
\end{equation}
Here T is a latent variable corresponding to semantic tree obtained from a semantic parser run on the question. W is the world, which is a representation of the image. This can be just the original image or the image along with additional features obtained from segmentation. $P(A=a|T,W)$ is evaluated using a deterministic evaluation function. $P(T|Q)$ is obtained by training a simple log-linear model. This model will be called SWQA (Single World Question Answering).

The authors further extend the SWQA model to a multi-world scenario to model uncertainty in segmentation and class labeling. Different labelings lead to different worlds, so the probability is now modeled as
\begin{equation}
\resizebox{0.5\textwidth}{!}{$P(A=a|Q, W) = \Sigma_{W}\Sigma_{T}P(A=a|T,W)P(W|S)P(T|Q)$}
\end{equation}
Here S is a set of segments along with a distribution over class labels for each segment. So sampling from the distribution for each segment will give us one possible world. As the above equation becomes intractable, the authors sample a fixed number of worlds from S. This model will be called MWQA (Multi World Question Answering).

These models are evaluated on the DAQUAR dataset.
\subsection{Non-attention Deep Learning Models}
Deep learning models for VQA typically involve the use of Convolutional Neural Networks (CNNs) to embed the image and word embeddings such as Word2Vec \cite{mikolov2013distributed} along with Recurrent Neural Networks (RNNs) to embed the question. These embeddings are combined and processed in various ways to obtain the answer. The following model descriptions assume that the reader is familiar with both CNNs \cite{krizhevsky2012imagenet} as well as RNN-variants like Long Short Term Memory units (LSTMs) \cite{hochreiter1997long} and Gated Recurrent Units (GRUs) \cite{cho2014learning}.

Some approaches do not involve the use of RNNs. These are discussed first. 
\subsubsection{iBOWIMG}
\cite{zhou2015simple} propose a baseline model called iBOWIMG for VQA. They use the output of a later layer of the pre-trained GoogLe Net model for image classification \cite{szegedy2015going} to extract image features. The word embeddings of each word in the question are taken as the text features, so the text features are simple bag-of-words. The image and text features are concatenated and softmax regression is performed across the answer classes. They showed that this model achieved performance comparable to several RNN based approaches on the VQA dataset.

\subsubsection{Full-CNN}
\cite{ma2015learning} propose a CNN-only model that we refer to here as Full-CNN. They use three different CNNs: an image CNN to encode the image, a question CNN to encode the question, and a join CNN to combine the image and question encoding together and produce a joint representation. 

The image CNN uses the same architecture as VGGnet \cite{simonyan2014very} and obtains a 4096-length vector from the second-last layer of this network. This is passed through another fully connected layer to get the image representation vector of size 400. The question CNN involves three layers of convolution + max pooling. The size of the convolutional receptive field is set to 3. In other words, the kernel looks at a word along with its immediate neighbors. The joint CNN, which they call the multi-modal CNN, performs convolution across the question representation with receptive field size 2. Each convolution operation is provided the full image representation. The final representation from the multi-modal CNN is given to a softmax layer to predict the answer. The model is evaluated on the DAQUAR and COCO-QA datasets.\\\\
The following models use both CNNs as well as RNNs.
\subsubsection{Ask Your Neurons (AYN) \cite{malinowski2016ask}}
This model uses a CNN to encode the image $x$ and obtain a continuous vector representation of the image. The question $q$ is encoded using an LSTM or a GRU network for which the input at time step $t$ is the word embedding for the $t^{th}$ question word $q_t$, as well as the encoded image vector. The hidden vector obtained at the final time step is the question encoding. A simple bag of words baseline the authors use is to encode the question as the sum of all the word embeddings.

The answer can be decoded in two different ways, either as a classification over different answers, or as a generation of the answer. Classification is performed by a fully connected layer followed by a softmax over possible answers. Generation, on the other hand, is performed by a decoder LSTM. This LSTM at each time point takes as input the previously generated word, as well as the question and image encoding. The next word is predicted using a softmax over the vocabulary. An important point to note is that this model shares some weights between the encoder and decoder LSTMs. The model is evaluated on the DAQUAR dataset.

\subsubsection{Vis+LSTM \cite{ren2015exploring}}
This model is very similar to the AYN model. The model uses the final layer of VGGnet to obtain the image encoding. They use an LSTM to encode the question. In contrast to the previous model, they provide the encoded image as the first `word' to this LSTM network, before the question. The output of this LSTM goes through a fully connected followed by softmax layer. They call this model Vis+LSTM.

The authors also propose a 2Vis+BLSTM model, which uses a bidirectional LSTM instead. The backward LSTM gets the image encoding as first input as well. The outputs of both LSTMs are concatenated and then passed through a fully connected and softmax layer.

\begin{table*}[]
\centering
\label{my-label}
\hspace*{-20mm}
\begin{tabular}{c|ccc|ccc|ccc}
                    & \multicolumn{3}{c|}{DAQUAR (Reduced)}                                                                                                                                                                      & \multicolumn{3}{c|}{DAQUAR (All)}                                                                                                                                                          & \multicolumn{3}{c}{COCO-QA}                                                                                                                                                    \\
\textbf{}           & \textbf{\begin{tabular}[c]{@{}c@{}}Accuracy\\ (\%)\end{tabular}} & \textbf{\begin{tabular}[c]{@{}c@{}}WUPS\\ at 0.9 (\%)\end{tabular}} & \textbf{\begin{tabular}[c]{@{}c@{}}WUPS\\ at 0 (\%)\end{tabular}} & \textbf{\begin{tabular}[c]{@{}c@{}}Accuracy\\ (\%)\end{tabular}} & \textbf{\begin{tabular}[c]{@{}c@{}}WUPS\\ at 0.9 (\%)\end{tabular}} & \textbf{\begin{tabular}[c]{@{}c@{}}WUPS \\ at 0 (\%)\end{tabular}} & \textbf{\begin{tabular}[c]{@{}c@{}}Accuracy\\ (\%)\end{tabular}} & \textbf{\begin{tabular}[c]{@{}c@{}}WUPS\\ at 0.9 (\%)\end{tabular}} & \textbf{\begin{tabular}[c]{@{}c@{}}WUPS\\ at 0 (\%)\end{tabular}} \\ \hline
\textbf{SWQA}       & 9.69                                                             & 14.73                                                               & 48.57                                                             & 7.86                                                             & 11.86                                                      & 38.79                                                     & -                                                       & -                                                          & -                                                        \\
\textbf{MWQA}       & 12.73                                                            & 18.10                                                               & 51.47                                                             & -                                                                & -                                                          & -                                                         & -                                                       & -                                                          & -                                                        \\
\textbf{Vis+LSTM}   & 34.41                                                            & 46.05                                                               & 82.23                                                             & -                                                                & -                                                          & -                                                         & 53.31                                                   & 63.91                                                      & 88.25                                                    \\
\textbf{AYN}        & 34.68                                                            & 40.76                                                               & 79.54                                                             & 21.67                                                            & 27.99                                                      & 65.11                                                     & -                                                       & -                                                          & -                                                        \\
\textbf{2Vis+BLSTM} & 35.78                                                            & 46.83                                                               & 82.15                                                             & -                                                                & -                                                          & -                                                         & 55.09                                                   & 65.34                                                      & 88.64                                                    \\
\textbf{Full-CNN}   & 42.76                                                            & 47.58                                                               & 82.60                                                             & 23.40                                                            & 29.59                                                      & 62.95                                                     & 54.95                                                   & 65.36                                                      & 88.58                                                    \\
\textbf{DPPnet}     & 44.48                                                            & 49.56                                                               & 83.95                                                             & 28.98                                                            & 34.80                                                      & 67.81                                                     & 61.19                                                   & 70.84                                                      & 90.61                                                    \\
\textbf{ATP}        & 45.17                                                            & 49.74                                                               & \textbf{85.13}                                                             & 28.96                                                            & 34.74                                                      & 67.33                                                     & 63.18                                                   & 73.14                                                      & 91.32                                                    \\
\textbf{SAN}                 & \textbf{45.50}                                                            & \textbf{50.20}                                                               & 83.60                                                             & \textbf{29.30}                                                            & \textbf{35.10}                                                      & \textbf{68.60}                                                     & 61.60                                                   & 71.60                                                      & 90.90                                                    \\
\textbf{CoAtt}               & -                                                                & -                                                                   & -                                                                 & -                                                                & -                                                          & -                                                         & 65.40                                                   & 75.10                                                      & 92.00                                                    \\
\textbf{AMA}                 & -                                                                & -                                                                   & -                                                                 & -                                                                & -                                                          & -                                                         & \textbf{69.73}                                                   & \textbf{77.14}                                                      & \textbf{92.50}                                                   
\end{tabular}
\caption{Results of various models on DAQUAR (reduced), DAQUAR (full), COCO-QA}
\end{table*}

\subsubsection{Dynamic Parameter Prediction (DPPnet) \cite{noh2016image}}
The authors of this paper argue that having a fixed set of parameters is not powerful enough for the VQA task. They take the architecture of VGG net, remove its final softmax layer and add three more fully connected layers, the last of which is followed by a softmax over possible answers. The second of these fully connected layers does not have a fixed set of parameters. Instead, the parameters come from a GRU network. This GRU network is used to encode the question, and the output from the network is passed through a fully connected layer to give a small vector of candidate parameter weights. This vector is then mapped to the larger vector of parameter weights required by the second fully connected layer above, using an inverse hashing function. This hashing technique is included by the authors to avoid having to predict the full set of parameter weights which could be expensive and may lead to over-fitting. The dynamic parameter layer can alternatively be seen as multiplying the image representation and question representation together to get a joint representation, as opposed to combining them in a linear fashion. The model is evaluated on the DAQUAR, COCO-QA and VQA datasets.

\subsection{Attention-based Deep Learning Techniques}
Attention based techniques are some of the most popular techniques that are being used across many tasks like machine translation \cite{bahdanau2014neural}, image captioning \cite{xu2015show} etc. For the VQA task, attention models involve focusing on important parts of the image, question or both in order to effectively give an answer.

\subsubsection{Where to Look}
\cite{shih2016look} propose an attention-based model henceforth referred to as WTL. They use VGGnet for encoding the image and concatenate the outputs of the final two layers of VGGnet to obtain image encoding. Question representation is obtained by averaging the word vectors of each word in the question. An attention vector is computed over the set of image features to decide which region in the image to give importance to. This vector is computed in the following way: If $V = (\overrightarrow{v_1}, \overrightarrow{v_2} \ldots \overrightarrow{v_K})$ is the set of image features, and $\overrightarrow{q}$ is the question embedding, then the importance of the $j^{th}$ region is computed as \\
$g_j = (A\overrightarrow{v_j}+b^A)^T(B\overrightarrow{q}+b^B)$\\
The attention weights are obtained on normalising $\overrightarrow{g}$. The final image representation is an attention weighted sum of the different regions. This is concatenated to the question embedding and passed to a dense+softmax layer.
The model  is evaluated on the VQA dataset. The loss function is a max margin based loss that takes into account the VQA evaluation metric.

\subsubsection{Recurrent Spatial Attention (R-SA) \cite{zhu2016visual7w}}
This model is a step above the previous model in two ways. Firstly, it uses LSTMs to encode the question, and secondly, it computes attention over the image repeatedly after scanning each word of the question. More concretely, we repeatedly compute an attention weighted sum of image features, $r_t$, at each time step $t$ of the LSTM. $r_t$ goes as an additional input to the next time step of LSTM. The attention weights $a_t$ used to obtain $r_t$ are computed using a dense+softmax layer over the previous hidden state of the LSTM $h_{t-1}$ and the image itself. Thus intuitively as we read the question we repeatedly decide which parts of the image to attend to, and parts to attend to now depend both on the current word as well as the previous attention weighted image through $h_{t-1}$.

This model is evaluated on the Visual7W dataset, for both the textual answering task as well as the pointing task (which points out a region in the image as an answer). The softmax cross-entropy loss between actual and predicted answer was used for the textual answering task. For the pointing task, log likelihood of a candidate region is obtained by taking dot product of feature representing that region and last state of LSTM. Again cross-entropy loss was used to train the model.

\begin{table*}[]
\centering
\label{my-label}
\begin{tabular}{c|ccccc|ccccc}
        & \multicolumn{5}{c|}{Test-Development}  & \multicolumn{5}{c}{Test-Standard}     \\
        & \multicolumn{4}{c}{Open Ended} & M.C. & \multicolumn{4}{c}{Open Ended} & M.C. \\
        & Y/N    & Number & Other & All   & All  & Y/N    & Number & Other & All   & All  \\ \hline
\textbf{iBOWIMG} & 76.5   & 35.0   & 42.6  & 55.7  & -    & 76.8   & 35.0   & 42.6  & 55.9  & -    \\
\textbf{DPPnet}  & 80.7   & 37.2   & 41.7  & 57.2  & -    & 80.3   & 36.9   & 42.2  & 57.4  & -    \\
\textbf{WTL}     & -      & -      & -     & -     & 62.4 & -      & -      & -     & -     & 62.4 \\
\textbf{AYN}     & 78.4   & 36.4   & 46.3  & 58.4  & -    & 78.2   & 36.3   & 46.3  & 58.4  & -    \\
\textbf{SAN}     & 79.3   & 36.6   & 46.1  & 58.7  & -    & -      & -      & -     & 58.9  & -    \\
\textbf{ATP}     & 80.5   & 37.5   & 46.7  & 59.6  & -    & 80.3   & \textbf{37.8}   & \textbf{47.6}  & 60.1  & -    \\
\textbf{NMN}     & \textbf{81.2}   & 38.0   & 44.0  & 58.6  & -    & \textbf{81.2}   & 37.7   & 44.0  & 58.7  & -    \\
\textbf{CoAtt}   & 79.7   & \textbf{38.7}   & \textbf{51.7}  & \textbf{61.8}  & \textbf{65.8} & -      & -      & -     & \textbf{62.1}  & \textbf{66.1} \\
\textbf{AMA}     & 81.01  &  38.42 & 45.23 & 59.17 & -    & 81.07  &  37.12 & 45.83 & 59.44 & -   
\end{tabular}
\caption{Results of various models on VQA dataset}
\end{table*}

\subsubsection{Stacked Attention Networks (SAN) \cite{yang2016stacked}}
This model is similar in spirit to the previous model in that it repeatedly computes attention over the image to get finer-grained visual information to predict the answer. However, while the previous model does this word by word, this model first encodes the entire question using either an LSTM or a CNN. This question encoding is used to attend over the image using a similar equation as before. Then the attention weighted image is concatenated with the question encoding and used to again compute attention over the original image. This can be repeated $k$ times after which the question and the final image representation are used to predict the answer. The authors argue that this sort of `stacked' attention helps the model to iteratively discard unimportant regions of the image. The authors experiment with k=1 and k=2 and report results on DAQUAR, COCO-QA and VQA datasets.

\subsubsection{Hierarchical Co-attention (CoAtt) \cite{lu2016hierarchical}}
This paper differs from the previous attention based methods in that in addition to modelling the visual attention, it also models question attention, that is, which part of the question to give importance to. They model two forms of co-attention: 1) Parallel co-attention, in which image and question attend over each other simultaneously. This is done by computing an affinity matrix $C = tanh(Q^TWI)$ where $W$ is a learnable weight matrix. $C_{ij}$ represents the affinity of the $i^{th}$ word in the question and $j^{th}$ region in the image. This matrix C is used to obtained both image and question attention vectors. 2) Alternating co-attention. In this we iteratively attend on image followed by query followed by image again (similar to SANs in spirit).

One additional idea that the authors use is encode the question at different levels of abstraction: word, phrase and question level. Question level representation is obtained by LSTM while word and phrase level representation are obtained from CNNs. They present results on VQA and COCO-QA datasets

\subsection{Other Models}
The following models use more ideas than simply changing how to attend to the image or question and as such do not fit in the previous sections.

\subsubsection{Neural Module Networks (NMNs) \cite{andreas2016deep}}
This model involves generating a neural network on the fly for each individual image and question. This is done through choosing from various sub-modules based on the question and composing these to generate the neural network. Modules are of five kinds: Attention[c] (which computes an attention map for a given image and given c; c can be `dog' for instance, then Attention[dog] will try to find a dog), classification[c] (which outputs a distribution over labels belonging to c for a given image and attention map; c can be `color'), reattention[c] (which takes an attention map and recomputes it based on c; c can be `above' which means shift attention upward), Measurement[c] (which outputs a distribution over labels based on attention map alone) and combination[c] (which merges two attention maps as specified by c; c could be `and' or `or').

To decide which modules to compose together, they first parse the question using a dependency parser and use this dependency to create a symbolic expression based on the head word. An example from the paper is `What is standing on the field?' becomes what(stand). These symbolic forms are then used to identify which modules to use. The whole system is then trained end to end through backpropagation. The authors test their model on the VQA dataset and also a more challenging synthetic dataset as they found that the VQA dataset did not require too much high level reasoning or composition.

\subsubsection{Incorporating Knowledge Bases}
\cite{wu2016ask} present the Ask Me Anything (AMA) model, that tries to leverage information from an external knowledge base to help guide visual question answering. It first obtains a set of attributes like object names, properties etc. of the images based on caption of the image. Image captioning model is trained on using standard image captioning techniques on the MS-COCO dataset. There are 256 possible attributes and the attribute generator is trained on MS-COCO using a variation of the VGG net. The top five attributes are used to generate queries for the DBpedia database \cite{auer2007dbpedia}. Each query returns a text which is summarized using Doc2Vec \cite{le2014distributed}. This summary is passed as an additional input to the decoder LSTM which generates the answer. The authors show results on VQA and COCO-QA datasets.

\section{Discussion and Future Work}
As has been the trend in recent years, deep learning models outperform earlier graphical model based approaches across all VQA datasets. However, it is interesting to note that the Answer Type Prediction (ATP) model performs better than the non-attention models, which proves that simply introducing convolutional and/or recurrent neural networks is not enough: identifying parts of the image that are relevant in a principled manner is important. ATP is even competitive with or better than some attention models like Where to Look (WTL) and Stacked Attention Networks (SAN). 

Significant improvement is shown by Hierarchical Co-Attention Networks (CoAtt), which was the first to attend on the question in addition to the image. This may be helpful especially for longer questions, which are harder to encode into a single vector representation by LSTMs/GRUs, so first encoding each word and then using the image to attend to important words helps the model perform better. The Neural Module Networks (NMN) uses the novel and interesting idea of automatically composing sub-modules for each image/question pair which performs similar to CoAtt on the VQA dataset, but outperforms all models on a synthetic dataset requiring more high level reasoning, indicating that this could be a valuable approach in the real world. However, more investigation is required to judge the performance of this model.

The best performing model on COCO-QA is Ask Me Anything (AMA) which incorporates information from an external knowledge base (DBpedia). A possible reason for improved performance is that the knowledge base helps answer questions that involve world or common sense knowledge that may not be present in the dataset. The performance of this model is not as good on VQA dataset, which might be because not too many questions in this dataset require world knowledge. This model naturally gives rise to two avenues for future work. The first would be recognizing when external knowledge is needed: some sort of model hybrid of CoAtt and AMA along with a decision maker for whether to access the KB might provide the best of both worlds. The decision might even be a soft one to enable end to end training. The second direction would be exploring the use of other knowledge bases like Freebase \cite{bollacker2008freebase}, NELL \cite{carlson-aaai} or OpenIE extractions \cite{schmitz2012open}.

As we have seen, novel ways of computing attention continue to improve performance on this task. This has been seen in the textual question answering task as well \cite{xiong2016dynamic} \cite{seo2016bidirectional}, so more recent models from that space can be used to guide VQA models. A study providing an estimated upper bound on performance for the various VQA datasets would be very valuable as well to get an idea for the scope of possible improvement, especially for COCO-QA which is automatically generated. Finally, most VQA tasks treat answering as a classification task. Only the VQA dataset allows for answer generation in a limited manner. It would be interesting to explore answering as a generation task more deeply, but dataset collection and effective evaluation methodologies for this remain an open question.

\section{Conclusion}
The field of VQA has grown by leaps and bounds despite being introduced just a few years ago. Deep learning methods for VQA continue to be the models receiving the most attention and showing state-of-the-art results. We surveyed the most prominent of these models and listed their performance over various large-scale datasets. Significant improvements in performance continue to be seen on many datasets, which means there is still plenty of room for future innovation in this task.

% include your own bib file like this:
%\bibliographystyle{acl}
%\bibliography{acl2017}
\bibliography{acl2017}
\bibliographystyle{acl_natbib}

\end{document}